\documentclass[sigconf]{acmart}

\copyrightyear{2022} 
\acmYear{2022} 
\setcopyright{acmcopyright}\acmConference[MM '22]{Proceedings of the 30th ACM International Conference on Multimedia}{October 10--14, 2022}{Lisboa, Portugal}
\acmBooktitle{Proceedings of the 30th ACM International Conference on Multimedia (MM '22), October 10--14, 2022, Lisboa, Portugal}
\acmPrice{15.00}
\acmDOI{10.1145/3503161.3547897}
\acmISBN{978-1-4503-9203-7/22/10}

\begin{document}

\title{Gait Recognition in the Wild with Multi-hop Temporal Switch}


\author{Jinkai Zheng}
\authornote{This work was done when Jinkai Zheng was an intern at JD Explore Academy.}
\email{zhengjinkai3@hdu.edu.cn}
\affiliation{%
  \institution{Hangzhou Dianzi University}
}

\author{Xinchen Liu}
\authornote{Corresponding author.}
\email{liuxinchen1@jd.com}
\affiliation{%
  \institution{JD Explore Academy}
}

\author{Xiaoyan Gu}
\email{guxiaoyan@iie.ac.cn}
\affiliation{%
  \institution{Institute of Information Engineering, Chinese Academy of Sciences}
}

\author{Yaoqi Sun}
\email{syq@hdu.edu.cn}
\affiliation{%
  \institution{Hangzhou Dianzi University}
}
\affiliation{%
  \institution{Lishui Institute of Hangzhou Dianzi University}
}

\author{Chuang Gan}
\email{ganchuang1990@gmail.com}
\affiliation{%
  \institution{MIT-IBM Watson AI Lab}
}

\author{Jiyong Zhang}
\authornotemark[2]
\email{jzhang@hdu.edu.cn}
\affiliation{%
  \institution{Hangzhou Dianzi University}
}

\author{Wu Liu}
\email{liuwu1@jd.com}
\affiliation{%
  \institution{JD Explore Academy}
}

\author{Chenggang Yan}
\authornote{The author is at the State Key Laboratory of Media Convergence Production Technology and Systems.}
\email{cgyan@hdu.edu.cn}
\affiliation{%
  \institution{Hangzhou Dianzi University}
}


\renewcommand{\shortauthors}{Jinkai Zheng et al.}
\begin{abstract}
Existing studies for gait recognition are dominated by in-the-lab scenarios.
Since people live in real-world senses, gait recognition in the wild is a more practical problem that has recently attracted the attention of the community of multimedia and computer vision.
Current methods that obtain state-of-the-art performance on in-the-lab benchmarks achieve much worse accuracy on the recently proposed in-the-wild datasets because these methods can hardly model the varied temporal dynamics of gait sequences in unconstrained scenes.
Therefore, this paper presents a novel multi-hop temporal switch method to achieve effective temporal modeling of gait patterns in real-world scenes.
Concretely, we design a novel gait recognition network, named Multi-hop Temporal Switch Network (MTSGait), to learn spatial features and multi-scale temporal features simultaneously.
Different from existing methods that use 3D convolutions for temporal modeling, our MTSGait models the temporal dynamics of gait sequences by 2D convolutions.
By this means, it achieves high efficiency with fewer model parameters and reduces the difficulty in optimization compared with 3D convolution-based models. 
Based on the specific design of the 2D convolution kernels, our method can eliminate the misalignment of features among adjacent frames. 
In addition, a new sampling strategy, i.e., non-cyclic continuous sampling, is proposed to make the model learn more robust temporal features.
Finally, the proposed method achieves superior performance on two public gait in-the-wild datasets, i.e., GREW and Gait3D, compared with state-of-the-art methods.
\end{abstract}

%


\maketitle



\begin{figure}[t]
  \centering
   \includegraphics[width=0.98\linewidth]{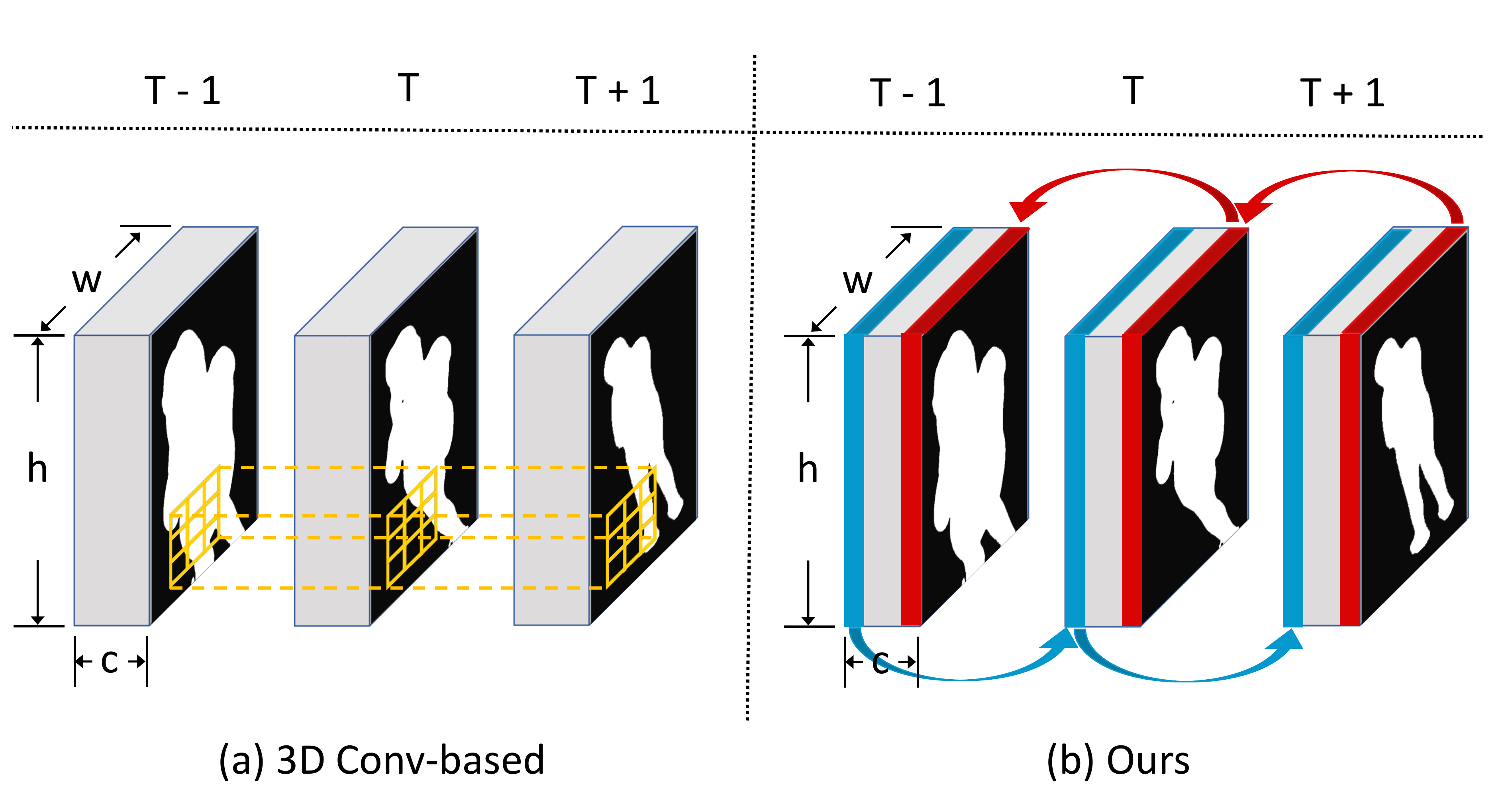}
   \caption{(a) The temporal modeling of 3D convolution-based methods by sliding spatial-temporal blocks over adjacent feature maps.
   (b) The temporal modeling of ours by switching channels between frames.
      T means time, h, w, and c represent the height, width, and channel dimensions of gait features, respectively. (Best viewed in color.)
   }
  \vspace{-3mm}
   \label{fig:figure_into}
\end{figure}

\section{Introduction}
Gait is a biological characteristic with great potential, which reflects the walking pattern of pedestrians.
Due to differences in movement and body shape, the target pedestrian can be uniquely identified by using gait~\cite{cvpr/NiyogiA94, csur/WanWP19}. 
Different from face and fingerprint, gait is remotely accessible, non-contacting, and hard to disguise, which makes it has unique potential in
social security.
However, gait recognition is still a very challenging task due to various uncertain factors in real-world scenes, such as occlusions, varied viewpoints, arbitrary walking styles, and so on~\cite{Zhu_2021_ICCV, zheng2022gait3d}. 

Commonly, according to the input types, gait recognition can be divided into two categories: model-based and model-free approaches. 
The model-based methods often take 2D/3D keypoints as inputs. 
However, due to the loss of much useful gait information like body shape, appearance, etc., the model-based methods are often inferior to model-free approaches in performance~\cite{zheng2022gait3d}. 
The model-free methods mainly use silhouettes as the gait representation. 
Recently, the methods based on deep learning achieve state-of-the-art performance on widely adopted gait recognition benchmarks like CASIA-B~\cite{icpr/YuTT06} and OU-MVLP~\cite{ipsjtcva/TakemuraMMEY18}. 
For example, 
GaitSet~\cite{aaai/ChaoHZF19} regarded a gait sequence as an unordered set and extracted the spatial-temporal information by max pooling, achieving the best performance at the time. 
But the temporal information is still lost. 
GaitGL~\cite{Lin_2021_ICCV} employed 3D convolution to extract spatial-temporal features and designed global and local branches to gather more useful gait knowledge, which significantly improve the performance of gait recognition. 
However, 3D convolution has the problem of feature misalignment, especially in the real-world scene, due to occlusion, arbitrary viewpoint, and other challenging factors, as shown in Fig.~\ref{fig:figure_into}. 

Although these methods have achieved excellent performance on the in-the-lab datasets, they cannot work well on the in-the-wild datasets like GREW~\cite{Zhu_2021_ICCV} and Gait3D~\cite{zheng2022gait3d}. 
We have carefully analyzed this phenomenon and found that the gait in the wild has many challenging factors, such as occlusions, varied viewpoints, arbitrary walking styles, and so on.
Existing temporal-based methods like GaitGL~\cite{Lin_2021_ICCV} and CSTL~\cite{Huang_2021_ICCV} did not fully consider the above challenges, so they obtained poor results on real-world gait datasets.
In particular, the existing methods of 3D convolution for temporal modeling still have the following three disadvantages: 
First, there is a problem of appearance misalignment in the operation of 3D convolution. 
In other words, the same position of the different frames may be different. 
As is shown in Fig.~\ref{fig:figure_into}, due to the challenges of 3D viewpoint and irregular walking style in real-world scenes, the position of the hand in the previous frame may be changed into the leg in the next frame.
Second, 3D convolution has a large number of parameters, so the optimization is particularly difficult.
Third, 3D convolution relies heavily on pre-training models, 
and then, as far as we know, there is no corresponding 3D pre-training model in the gait recognition field.

To address the above-mentioned shortcomings and to better model the temporal information of gait recognition in the wild, this paper proposes a multi-hop temporal switch approach. 
As is shown in Fig.~\ref{fig:figure_into}, to avoid the feature misalignment problem in 3D convolution-based methods, we adopt channel switching between frames to achieve temporal modeling. 
The experimental results and visualization prove that our method does make the model pay more attention to the motion information, such as foot movement, 
In addition, we propose a novel gait recognition network, i.e., Multi-hop Temporal Switch Network (MTSGait). 
The MTSGait is composed of two branches: the spatial branch and the multi-hop temporal switch branch.
The spatial branch can reduce the damage to appearance features caused by switch operation. 
The multi-hop temporal switch branch allows the model to learn multi-scale temporal information. 
It is important to point out that all of the operations in our method are relying on 2D convolution. 
That is to say, our method does not have the problems of large-scale model parameters and difficulty in training. 
Moreover, 
We find that the previous assumption of gait as a cycle motion is problematic, so we propose a novel gait data sampling strategy, i.e., Non-cyclic continuous sampling, which can make the model learn more robust temporal features.

In summary, the contributions of this paper are as follows:
\begin{itemize}
  \item We propose a novel gait recognition network, i.e., MTSGait. The network is composed of the spatial branch and multi-hop temporal switch branch, which guarantees the model can learn spatial and multi-scale temporal information simultaneously.
  \item Our MTSGait realizes the temporal modeling by 2D convolution without the problem of large-scale model parameters and difficulty in training like 3D convolution-based approaches.
  \item To better extract temporal information on gait recognition, especially in real-world scenes, we propose a new sampling strategy, i.e., Non-cyclic continuous sampling, which can improve the model to learn more robust temporal features.
  \item Our method achieves superior performance on two public gait in-the-wild datasets, i.e., GREW and Gait3D, compared with state-of-the-art methods.
\end{itemize}

\section{related work}
\subsection{Gait Recognition}
At present, there are two main categories of gait recognition methods: model-based and model-free approaches~\cite{csur/WanWP19, corr/scf_survey}. 
\textbf{Model-based} methods focus on structural modeling of the human body, such as 2D/3D human keypoints, etc. 
Early methods mainly belong to the model-based. 
For example, Yam~\textit{et al.}~\cite{pr/YamNC04} employed the coupled oscillators and the biomechanics of human locomotion to model walking and ruining patterns. 
Yamauchi~\textit{et al.}~\cite{cvpr/YamauchiBS09} proposed the first method using 3D keypoints from RGB frames for walking human recognition.
Ariyanto and Nixon~\cite{icb/AriyantoN11} applied a complex multi-camera system to build a 3D voxel-based dataset and proposed a structural model of articulated cylinders with 3D Degrees of Freedom at each joint to model the human lower legs.
Recently, there are also some model-based gait recognition methods. 
Liao~\textit{et al.}~\cite{pr/LiaoYAH20} defined joint angle, limb length, and joint motion based on the 3D keypoints, and used the three defined information combined with pose features as the gait representation. 
Teepe~\textit{et al.}~\cite{corr/abs-2101-11228} modeled the 2D skeleton as a graph and adopted a Graph Convolution Network, i.e., the ResGCN~\cite{mm/Song0SW20}, to learn features by the contrastive loss. 
Li~\textit{et al.}~\cite{accv/LiMXYYR20} combined both 2D and 3D keypoints information of the human body as gait representation, and use Convolutional Neural Networks (CNNs)~\cite{nips/KrizhevskySH12} to extract gait features. 

The model-based approaches are robust to variants of clothing and camera viewpoints. 
However, due to the loss of much useful gait information like body shape information, model-based methods are often inferior to model-free approaches in performance. 
\textbf{Model-free} methods mainly use the silhouettes as the gait representation. 
In early times,  Han~\textit{et al.}~\cite{pami/HanB06} proposed to compress the silhouettes of the sequence in the temporal dimension, thus obtaining Gait Energy Image (GEI).
Recently, due to the success of deep learning on multimedia and computer
vision tasks~\cite{pami/cgy1, tomccap/cgy2, tcsv/cgy3, tomm/cgy4, tomm/cgy5, liuwu1, liuwu2, lxc2, xuhang1, Liu_tc3d}, deep learning-based methods also dominated the performance of gait recognition.
For example, Shiraga~\textit{et al.}~\cite{icb/ShiragaMMEY16} and Wu~\textit{et al.}~\cite{pami/WuHWWT17} proposed to learn gait features from GEIs by CNNs and significantly outperformed previous methods. 
Zhang~\textit{et al.}~\cite{liuwu_gait1} developed a Siamese neural network-based gait recognition framework to automatically extract robust and discriminative gait features from GEIs.
The most recent methods started to extract gait features directly from silhouette sequences. 
For example, Chao~\textit{et al.}~\cite{aaai/ChaoHZF19} regarded the gait sequence as a set regardless of temporal information and used CNNs to extract the frame-level and set-level features. 
Hou~\textit{et al.}~\cite{eccv/HouCLH20} designed a compact block that reduced the dimension of the gait features from 15,872 to 256. 
Fan~\textit{et al.}~\cite{cvpr/FanPC0HCHLH20} proposed a novel network, named GaitPart, to split the gait silhouettes horizontally and extract detailed features from each part. 
Huang~\textit{et al.}~\cite{iccv/HuangXS0LH021} proposed a 3DLocalCNN network, which can achieve the location and feature extraction of more refined human body parts. 
Lin~\textit{et al.}~\cite{Lin_2021_ICCV} used 3D convolution to extract spatial and temporal information at the same time, and proposed a GLConv module to aggregate the global and local features. 
Zheng~\textit{et al.}~\cite{zjk_gait1} made one of the first explorations for unsupervised cross-domain gait recognition with a Transferable Neighborhood Discovery framework.
Although these methods have greatly improved the accuracy of gait recognition, they are only carried out in the lab senses, i.e., constrained environments. 
People live in real-world senses, i.e., unconstrained environments. 
It is important to promote gait recognition in in-the-wild scenarios. 

Fortunately, some researchers have started to focus on this field and have produced large-scale gait benchmarks in real scenarios, which promote gait recognition from experimental research to practical application. 
Zhu~\textit{et al.}~\cite{Zhu_2021_ICCV} first built one of the largest gait in the wild benchmark named GREW. 
It contains 26K subjects and 128K sequences with rich attributes from 882 cameras in a large public area, which makes it the first dataset for unconstrained gait recognition. 
They also performed extensive gait recognition experiments, including representative methods, attributes analysis, etc. 
The results showed that gait recognition in the wild is a very challenging task for current SOTA methods.
Zheng~\textit{et al.}~\cite{zheng2022gait3d} pointed out that humans live and walk in the unconstrained 3D space, so projecting the 3D human body onto the 2D plane will discard a lot of crucial information like the viewpoint, shape, etc. 
To realize 3D gait recognition in the real-world scenario, they built the first large-scale 3D gait recognition dataset, named Gait3D, which provides the 3D human meshes, 2D/3D keypoints, and 2D silhouettes of gait collected from unconstrained environments. 
They also proposed a novel 3D gait recognition framework, named SMPLGait, to explore 3D human meshes for gait recognition. 
Based on the Gait3D dataset, they did extensive experiments and found that the existing SOTA methods usually fail on Gait3D, even though the excellent performance on in-the-lab datasets, e.g., CASIA-B~\cite{icpr/YuTT06} and OU-LP~\cite{tifs/IwamaOMY12}.
The above two benchmarks play a great role in promoting gait recognition in real-world scenes. 
To promote the application of gait recognition, this paper mainly focuses on the task of gait recognition in the wild. 

\subsection{Temporal Modeling}
Early researchers artificially constructed temporal information for gait recognition. 
For example, Urtasun and Fua~\cite{fgr/UrtasunF04} employed the 3D temporal motion models using an articulated skeleton for gait analysis. 
Zhao~\textit{et al.}~\cite{fgr/ZhaoLLP06} proposed a local optimization algorithm to track 3D motion for gait recognition. 
Recent gait temporal modeling methods can be divided into four categories: Set-based, LSTM-based, 1DCNN-based, and 3DCNN-based. 
GaitSet~\cite{aaai/ChaoHZF19} and GLN~\cite{eccv/HouCLH20} regarded a gait sequence as an unordered set and extracted the spatio-temporal information by temporal pooling. 
However, this loses a lot of useful temporal features. 
Zhang~\textit{et al.}~\cite{cvpr/ZhangT0A0WW19} and Huang~\textit{et al.}~\cite{tip/ZhangHYW20} applied LSTM to achieve short-long temporal modeling. 
GaitPart~\cite{cvpr/FanPC0HCHLH20} and CSTL~\cite{Huang_2021_ICCV} employed 1D convolution to model short-term and long-term temporal features. 
MT3D~\cite{mm/Lin_mt3d} and GaitGL~\cite{Lin_2021_ICCV} utilized 3D convolution to model spatial-temporal features directly, but these methods are often difficult to train and expensive in computational memory. 
Lin~\cite{iccv/LinGH19} proposed that temporal modeling can be achieved by sliding channel information over timing. 
However, we found that directly transferring channel information may harm spatial features, which can decrease the performance of gait recognition. 
In this paper, we build a novel gait recognition network, i.e., MTSGait, which contains the spatial branch and multi-hop temporal switch branch. 
The structure guarantees the model can learn spatial and multi-scale temporal knowledge simultaneously. 
In addition, the proposed MTSGait network is completely based on 2D convolution without the problem of large-scale model parameters and difficulty in training. 
\begin{figure*}[t]
  \centering
   \includegraphics[width=0.98\linewidth]{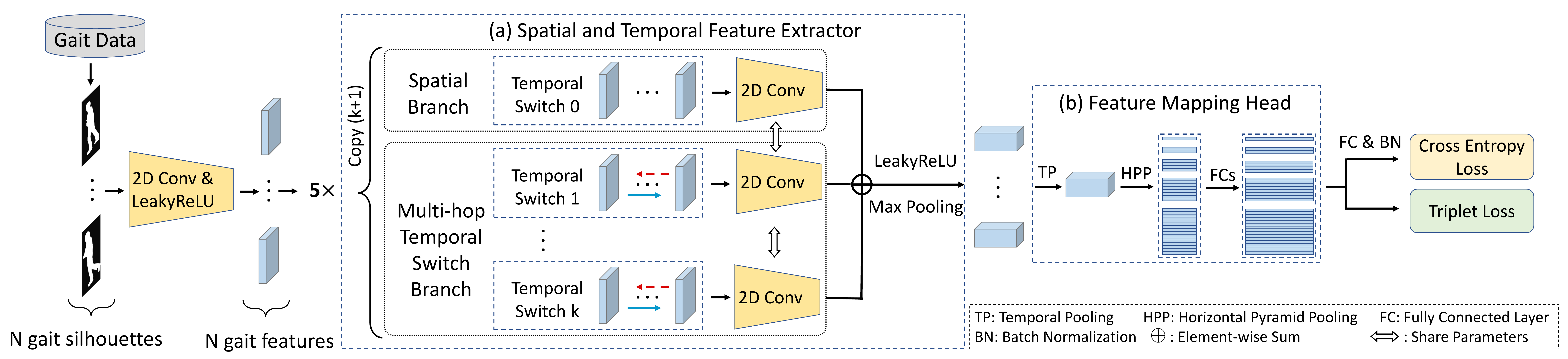}
   \caption{Overview of the proposed MTSGait framework. (Best viewed in color.)}
   \label{fig:figure_framework}
\end{figure*}

\section{The Proposed Method}
In this section, we first introduce the overview of the proposed framework. 
Then, we describe the key component of our method, including the Spatial and Temporal Feature Extractor and Multi-hop Temporal Switch (MTS).
Next, we discuss the existing gait data sampling strategy and propose a new sampling method.
At last, the details of training and inference are presented.

\subsection{Overview}
The overview of the proposed gait recognition framework is shown in Fig.~\ref{fig:figure_framework}. 
First, the sampled sequences of silhouettes are fed into a 2D convolution to extract the shallow features.
We formulate $X_{i}= \{ \mathbf{x}_i\}^{N}_{i=1}$ as the input sequence, where $x_{i}\in \mathbb{R}^{1 \times H \times W}$, $N$ is the length of the sampled sequence, $H$ and $W$ are the height and width of the gait silhouette. The process can be formulated as:
\begin{equation}
    \mathbf{F}_i = ReLU(f_{b}^{a \times a}(X_i)),
\label{equ:2DConv}
\end{equation}
where $ReLU(\cdot)$ is the LeakyReLU activation function, $f_b^{a \times a}(\cdot)$ is the 2D convolution with kernel size $a$ and stride $b$ aims to extract frame-level features from each gait silhouette, $\mathbf{F}_i \in \mathbb{R}^{N \times c \times h \times w}$ is the feature map for $\mathbf{X}_i$,
$c$ is the number of channels, $h$ and $w$ are the height and width of the feature maps.

Next, the Spatial and Temporal Feature Extractor is designed to integrate spatial and temporal information.
Then, we introduce a feature mapping head to further map the features. 
At last, we use triplet loss and cross-entropy loss to train the model.
We will introduce the above modules in detail.

\subsection{Spatial and Temporal Feature Extractor}
As is shown in Fig.~\ref{fig:figure_framework}, the Spatial and Temporal Feature Extractor consists of the spatial branch, the multi-hop temporal switch (MTS) branch, LeakyReLU, and max pooling. 
The MTS branch can also be divided into different temporal scale branches, which can help the model learn more temporal information from the gait sequence.
Due to our temporal branch may harm the spatial features, the spatial branch is necessary to be added. 
Specifically, the spatial branch is implemented by 2D convolution.
The process can be formulated as follows:
\begin{equation}
    \mathbf{G}_i = MTS(\mathbf{F}_i) + f_{b}^{a \times a}(\mathbf{F}_i),
\label{equ:STFE}
\end{equation}
where $MTS(\cdot)$ is the operation of MTS branch, $\mathbf{G}_i \in \mathbb{R}^{N \times c^{\prime} \times h^{\prime} \times w^{\prime}}$ is the feature map containing spatial and multi-scale temporal information. 

\begin{figure}[t]
  \centering
   \includegraphics[width=0.98\linewidth]{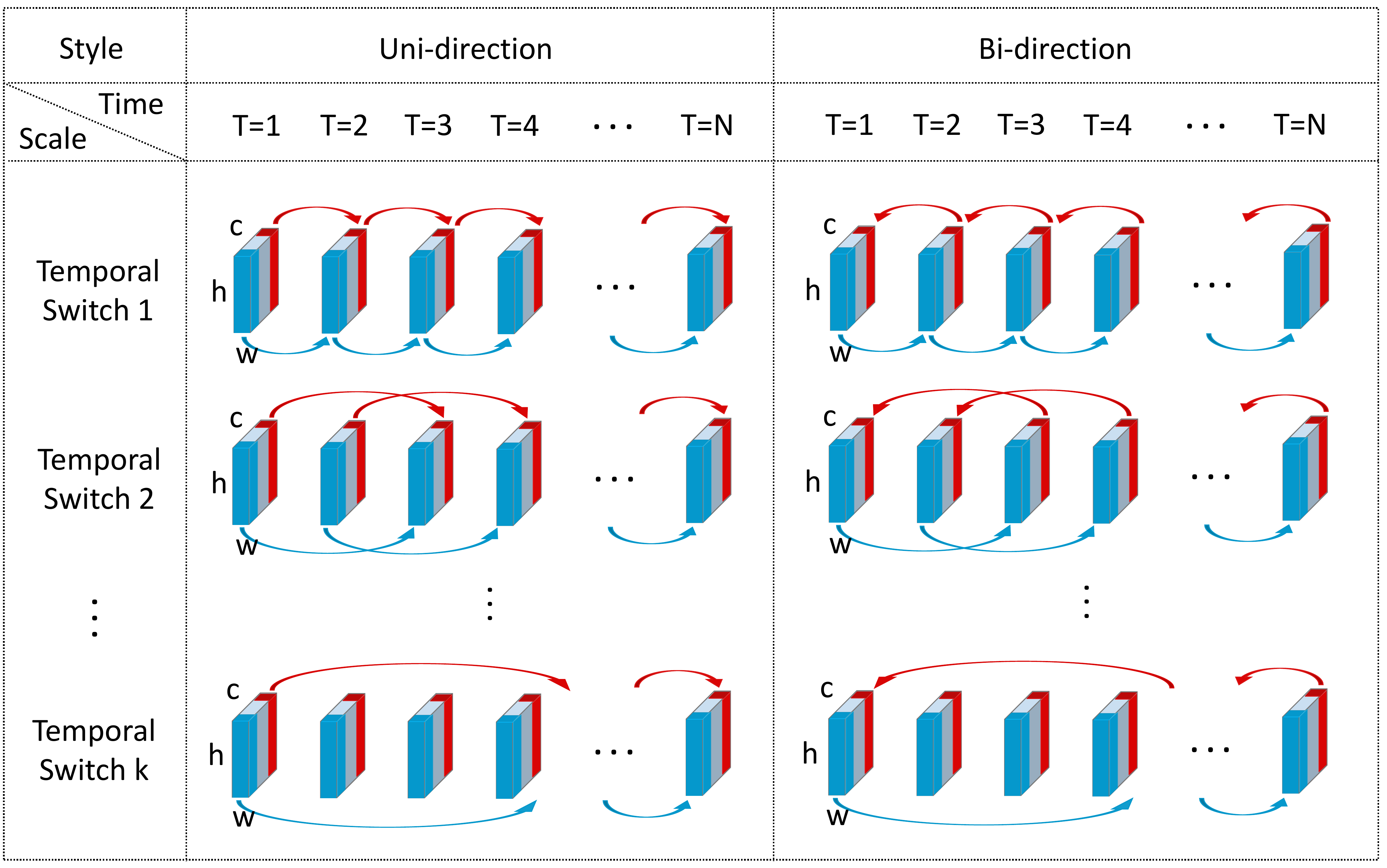}
   \caption{Illustration of the uni-direction, bi-direction, and multi-hop temporal switch operation. (Best viewed in color.) }
   \label{fig:figure_mtt}
\end{figure}

\subsection{Multi-hop Temporal Switch (MTS)}
Recently, many temporal modeling methods, e.g., MT3D~\cite{mm/Lin_mt3d} and GaitGL~\cite{Lin_2021_ICCV} have been proposed to extract temporal information by using 3D convolution.
However, due to the challenges of 3D viewpoint and irregular walking style in real-world scenes, the position of the hand may become a leg in the next frame.
The 3D convolution works by sliding spatial-temporal blocks over adjacent feature maps. 
This working mechanism makes its features in the temporal level seriously misaligned especially in the real-world sense, as is shown in Fig.~\ref{fig:figure_into}. 
In addition, the 3D convolution has a large number of parameters, so optimization is particularly difficult.
That is, training 3D convolution relies heavily on pre-training models, 
As far as we know, there is no corresponding 3D pre-training model in the gait recognition field.

To solve the above problems, we introduce the MTS branch to conduct multi-scale temporal modeling. 
Based on the assumption that 2D convolution kernels are particularly specific to different patterns~\cite{cvpr/ZhangWZ18a}, we switch channels between frames to realize the temporal modeling. 
Suppose the weight of the 2D convolution is split into three parts as $W = (\omega_1, \omega_2, \omega_3)$, and the input feature map of frame $t$ can be split in the channel dimension as $F_{t} = (F_t^{1}, F_t^{2}, F_t^{3})$. 
The operation of 2D convolution on frame $T$ can be written as:
\begin{equation}
    f_{b}^{a \times a}(\mathbf{F}_t) = \omega_1 F_t^{1} + \omega_2 F_t^{2} + \omega_3 F_t^{3},
\label{equ:origin_2Dconv}
\end{equation}
Now, we switch part of channels, i.e., $F_t^{1}$ is replaced with $F_{t-1}^{1}$, $F_t^{3}$ is replaced with $F_{t+1}^{3}$, $F_t^{2}$ is left unchanged. 
The operation can be represented like this:
\begin{equation}
    f_{b}^{a \times a}(\mathbf{F}_{t-1}, \mathbf{F}_{t}, \mathbf{F}_{t+1}) = \omega_1 F_{t-1}^{1} + \omega_2 F_t^{2} + \omega_3 F_{t+1}^{3},
\label{equ:channle_switch_bi_1}
\end{equation}
In this way, the 2D convolution operation can integrate the information about the previous frame, the current frame, and the next frame. 
We can also conduct a longer time span version. 
The process can be formulated as:
\begin{equation}
    f_{b}^{a \times a}(\mathbf{F}_{t-j}, \mathbf{F}_{t}, \mathbf{F}_{t+j}) = \omega_1 F_{t-j}^{1} + \omega_2 F_t^{2} + \omega_3 F_{t+j}^{3},
\label{equ:channle_switch_bi_i}
\end{equation}
At last, we combine the multi-hop temporal switch features, which realizes the multi-scale temporal modeling. 
The operation can be represented like this:
\begin{equation}
    MTS(\mathbf{F}_i) = \sum_{j=1}^{S} f^{a \times a}_{b}(\mathbf{F}_{i-j}, \mathbf{F}_{i}, \mathbf{F}_{i+j}), j=1,2,\dots,S,
\label{equ:mts}
\end{equation}
where $S$ is the time span, $\mathbf{F}_i$ is the input feature map of the MTS module. 
In our implementation, we divide the feature maps into $m$ parts in the channel dimension, and the features of the first and last parts are switched in uni-direction or bi-direction ways, as is shown in Fig.~\ref{fig:figure_mtt}.


\subsection{Feature Mapping Head}
\label{subsec:feature_mapping}
Since the length of the input silhouette sequence may be different, we employ temporal pooling to aggregate the temporal-level features~\cite{aaai/ChaoHZF19}. 
Assume that $X_{in} \in \mathbb{R}^{B \times N \times C_{in} \times H_{in} \times W_{in}}$ is the input of the feature mapping head, where $B$ is the batch size, $N$ is the length of gait sequence, $C_{in}$ is the number of input channels, $H_{in}$ and $W_{in}$ are the height and width of the feature maps. 
The temporal pooling can be formulated as:
\begin{equation}
    X_{out} = Maxpooling_{t}(X_{in}),
\label{equ:temporal_pooling}
\end{equation}
where $X_{out} \in \mathbb{R}^{B \times C_{in} \times H_{in} \times W_{in}}$ is the sequence-level feature and $Maxpooling_{t}(\cdot)$ is the max pooling layer. 

After temporal pooling, the dimensions of feature maps are unified. 
To obtain more discriminative features, we implement horizontal pyramid pooling to generate the local information for each horizontal strip. 
The feature maps are first split into strips, and then max pooling and mean pooling are used to extract refined features from these strips. 
At last, we employ multiple separate fully connected layers (FCs) to further aggregate more discriminative information. 
The process can be presented as follows:
\begin{equation}
    Y_{out} = FCs(Maxpooling(X_{out}) +Meanpooling(X_{out})).
\label{equ:temporal_pooling}
\end{equation}

\subsection{Gait Sampling Strategy}
Existing video-based person re-identification and sequence-based gait recognition methods sample a sequence with a fixed length as the input. 
Chao~\textit{et al.}~\cite{aaai/ChaoHZF19} proposed the idea of regarding gait sequence as a set, thereby discarding the order between frames and randomly sampling N frames from a gait sequence.
Recent studies have advocated exploring the temporal information of gait.
Hou~\textit{et al.}~\cite{cvpr/HouCM0S21_video_reid} orderly sample frames with a stride of fixed length, i.e., uniform sampling. 
Based on the assumption that gait is a periodic motion, Lin~\textit{et al.}~\cite{Lin_2021_ICCV} and Fan~\textit{et al.}~\cite{cvpr/FanPC0HCHLH20} propose a cyclic continuous sampling strategy. 
Specifically, the sample frames of fixed length consecutively in a gait sequence, and when the sequence length is short to the sample length, they loop through the sequence, completing it until the sample length is reached.

However, we find that there is often a great difference between the beginning and the end of a gait sequence, which is a great noise to the model. 
Especially when the dataset has a large number of gait sequences whose length is less than the sampling length, this noise is fatal for the model to learn temporal knowledge.
Based on the above analysis, we designed a new sampling Strategy, i.e., Non-cyclic continuous sampling. 
We formulate $X = \{ \mathbf{x}_1, {x}_2, ..., {x}_L\}$ as the input sequence, where $ \mathbf{x}_i \in \mathbb{R}^{H \times W}$ is the $i$-th binary frame, $L$ is the length of the sequence, $H$ and $W$ are the height and width of the silhouette image.
The process of Non-cyclic continuous sampling can be formulated as:
\begin{equation}
\begin{aligned}
    X_{N} &= \begin{cases}{
                    \left[x_{(t+1)}, x_{(t+2)}, \ldots, x_{(t+N)}\right],} & L>=N, \\
                    \operatorname{Sort}\left(\left[x_{1}, x_{2}, \ldots, x_{L}, x_{1}, \ldots, x_{(N-L)}\right]\right), & L<N,
             \end{cases} \\
         &= \begin{cases}{
                    \left[x_{(t+1)}, x_{(t+2)}, \ldots, x_{(t+N)}\right],} & L>=N, \\
                    {\left[x_{1}, x_{1}, \ldots, x_{(N-L)}, x_{(N-L)}, x_{(N-L+1)}, \ldots, x_{L}\right],} & L<N,
             \end{cases}
\end{aligned}
\label{equ:eq_sampling}
\end{equation}
where $N$ is the sampling length, $ t \in [0, L-N]$ is the random sampled start index, $X_N$ is the sampled sequence. 
The experiments in Sec~\ref{subsubsec:sampling_ablation_study} verify that the proposed Non-cyclic continuous sampling strategy can make the model learn more robust temporal information.

\subsection{Training and Inference}
\textbf{Training.} Like the loss function commonly used in person re-identification~\cite{corr/HermansBL17_triplet_loss} and gait recognition~\cite{mm/Lin_mt3d}, we also adopt the Batch ALL triplet loss~\cite{corr/HermansBL17_triplet_loss} and cross-entropy loss to train our model. 
The triplet loss can decrease the intra-class distance and increase the inter-class distance. 
The final loss can be defined as:
\begin{equation}
    L_{final} = \alpha L_{tri} + \beta L_{ce},
\label{equ:final_loss}
\end{equation}
where $\alpha$ and $\beta$ are the weighting parameters, $L_{tri}$ is the triplet loss, $L_{ce}$ is the cross entropy loss. 
The triplet loss can be defined as:
\begin{equation}
L_{tri} = \frac{1}{N_t} \sum_{i=1}^{N_t} [||F(X_i^a)-F(X_i^p)|| - ||F(X_i^a)-F(X_i^n)|| + m]_{+},
\label{equ:triplet_loss}
\end{equation}
where $N_t$ is the number of triplets of non-zero
loss terms in a mini-batch, $F(\cdot)$ denote the proposed model, $X_i^a$ is the anchor sequence. $X_i^p$ and $X_i^n$ are positive and negative sequences with respect to the anchor, respectively. 
$\|\cdot\|$ is the euclidean distance, $m$ is the margin parameter, and $[\gamma]_{+}$ means $max(\gamma, 0)$.

The batch size for training is $p \times k$, where $p$ denotes the number of training subjects, and $k$ denotes the number of training samples corresponding to each subject.
In particular, we calculate the triplet loss for each horizontal strip feature obtained in Sec~\ref{subsec:feature_mapping} separately. 
In the training phase, considering the limitations of memory, we fix the length of each gait sequence. 

\textbf{Inference.} In the inference phase, to extract the complete gait information as much as possible, we orderly load all the frames of each gait sequence. 
According to the official manner commonly used in gait recognition, the test set is divided into the probe set $\mathbf{Q}$ and the gallery set $\mathbf{G}$. 
The gait sequence in  the probe set is the query sample, and the gait sequence in the gallery set is used to be retrieved. 
The common metric strategy is to calculate the euclidean distance or cosine distance between samples from $\mathbf{Q}$ and $\mathbf{G}$. 
For a fair comparison with other methods, we utilize euclidean distance on GREW dataset and cosine distance on the Gait3D dataset. 

\section{experiments}
\subsection{Datasets and Protocols}
We evaluate the proposed method on two commonly used gait in-the-wild datasets, i.e., Gait3D~\cite{zheng2022gait3d} and GREW~\cite{Zhu_2021_ICCV}. 
The details of the two datasets and protocols are as follows:

\textbf{Gait3D.} 
The Gait3D dataset~\cite{zheng2022gait3d} is a newly proposed gait in-the-wild dataset. 
It includes 4,000 subjects, 25,309 sequences, and 3,279,239 frame images in total, which were extracted from 39 cameras in an unconstrained indoor scene, i.e., a large supermarket. 
Following the official train/test strategy~\cite{zheng2022gait3d}, 3,000 subjects are selected as the training set and another 1,000 subjects as the testing set. 
For each subject in the testing set, one sequence is registered as the query, and the rest of the sequences become the gallery. 
This dataset has the following special factors: 3D viewpoint, irregular walking speed, occlusion, etc. 
All these factors make gait recognition in the wild a great challenge task. 
The evaluation protocol is based on the open-set instance retrieval setting. 
We adopt the average Rank-1 and Rank-5 accuracy over all query sequences.
We also adopt the mean Average Precision (mAP) and mean Inverse Negative Penalty (mINP)~\cite{corr/abs-2001-04193} which consider the recall of multiple instances and hard samples.

\textbf{GREW.} 
The GREW dataset~\cite{Zhu_2021_ICCV} is one of the largest gait in-the-wild datasets. 
It contains 26,345 subjects from 882 cameras in a large public area. 
We follow the evaluation protocol proposed in ~\cite{Zhu_2021_ICCV}. 
There are 20,000 subjects for training and 6,000 subjects for testing. 
In the testing set, there are 4 sequences in each subject, and 2 for the query set and 2 for the gallery set. 
To make it more suitable for real-world applications, GREW is also enriched by a distractor set with 233K gait sequences. 
However, to make a fair comparison with other methods, the distractor set is removed in this work. 
We adopt the Rank-1, Rank-5, Rank-10, and Rank-20 accuracy to evaluate the performance of GREW dataset. 


\subsection{Implementation Details}
Following the same processing approach as ~\cite{aaai/ChaoHZF19}, we normalize the image of each frame to $64 \times 44$ for the Gait3D dataset. 
For the GREW dataset, the original image size is normalized to $64 \times 44$ as well. 
The channel number of the first 2D convolution layer and the subsequent five layers is set to 64, 64, 128, 128, 256, and 256, respectively. 
The kernel size is set to 3, except for the first 2D convolution layer which is set to 5. 
The padding is set to 1, except for the first 2D convolution layer which is set to 2. 
The moving stride is set to 1. 
For GREW dataset, since it contains 5 times more sequences
than Gait3D, we add two additional 32-channel convolution layers at the front of the network. 
It is worth noting that both the spatial branch and the multi-hop temporal switch branch contained in the MTS module share parameters of the same 2D convolution. 
$m$ in Equ.~\ref{equ:triplet_loss} is set to 0.2. 
$\alpha$ in Equ.~\ref{equ:final_loss} is set to 1.0. 
$\beta$ in Equ.~\ref{equ:final_loss} is set to 0.1. 
For the Gait3D dataset, we follow the settings as ~\cite{zheng2022gait3d}. 
The batch size is $ p \times k = 32 \times 4$, where 32 is the number of subjects, and 4 is the number of gait sequences per subject. 
Considering the expensive computation of memory during the training phase, we set the number of frames $N$ to 30. 

During the test phase, all frames of each gait sequence are utilized, and the maximum number of frames per sequence is limited to 720, also for memory consideration. 
For the Gait3D dataset, the model is trained for 180K iterations with the initial Learning Rate (LR)=1e-3, and the LR is multiplied by 0.1 at the 30K and 90K iterations.
Adam~\cite{corr/KingmaB14} is taken as the optimizer and the weight decay is set to 5e-4. 
For the GREW dataset, most of the settings follow those of Gait3D. 
However, to adapt to the large scale of GREW, we make the following adjustments. 
The model is trained for 300K iterations with the initial LR=1e-2 and the LR is multiplied by 0.1 at the 150K and 250K iterations. 
In addition, SGD~\cite{corr/Ruder16} is taken as the optimizer and the weight decay is set to 5e-4.

\begin{table}[t]
\centering
\begin{tabular}{l|cccc}
Methods                      			& Rank-1 & Rank-5 & mAP & mINP \\ \midrule[1.5pt]
PoseGait~\cite{pr/LiaoYAH20}	        & 0.24  & 1.08  & 0.47  & 0.34 \\ 
GaitGraph~\cite{corr/abs-2101-11228}	& 6.25  & 16.23 & 5.18  & 2.42 \\  \midrule
GEINet~\cite{icb/ShiragaMMEY16}      	& 5.40  & 14.20 & 5.06  & 3.14 \\
GaitSet~\cite{aaai/ChaoHZF19}		    & 36.70 & 58.30 & 30.01 & 17.30	\\
GaitPart~\cite{cvpr/FanPC0HCHLH20}		& 28.20 & 47.60 & 21.58 & 12.36	\\ 
GLN~\cite{eccv/HouCLH20}				& 31.40 & 52.90 & 24.74 & 13.58	\\
GaitGL~\cite{Lin_2021_ICCV}	            & 29.70 & 48.50 & 22.29 & 13.26	\\ 
CSTL~\cite{Huang_2021_ICCV}	            & 11.70 & 19.20 & 5.59  & 2.59	\\ 
SMPLGait~\cite{zheng2022gait3d}         & 46.30 & 64.50 & 37.16 & \textbf{22.23}  \\  \midrule
Ours  w/o MTS	                        & 42.90 & 63.90 & 35.19 & 20.83	\\ 
Ours                                    & \textbf{48.70}	& \textbf{67.10} & \textbf{37.63} & 21.92  \\  \end{tabular} 
\caption{Comparison of the state-of-the-art gait recognition methods on Gait3D.} 
\vspace{-7mm}
\label{tab:experiments_gait3d}
\end{table}

\subsection{Experimental results on Gait3D}
In this section, we compare the proposed method with several state-of-the-art methods, including PoseGait~\cite{pr/LiaoYAH20}, GaitGraph~\cite{corr/abs-2101-11228}, GEINet~\cite{icb/ShiragaMMEY16}, GaitSet~\cite{aaai/ChaoHZF19}, GaitPart~\cite{cvpr/FanPC0HCHLH20}, GLN~\cite{eccv/HouCLH20}, GaitGL~\cite{Lin_2021_ICCV}, CSTL~\cite{Huang_2021_ICCV}, and SMPLGait~\cite{zheng2022gait3d} on Gait3D dataset. 
As is shown in Table~\ref{tab:experiments_gait3d}, model-based methods like PoseGait and GaitGraph are greatly worse than model-free methods. 
This indicates that it is difficult to perform gait recognition in real-world scenes only by relying on a few sparse human body joints. 
Among the model-free approaches, GEINet gets the worst results, indicating that GEIs also discard a lot of useful gait information. 
Moreover, those approaches that consider temporal modeling, such as GartPart, GaitGL, and CSTL, achieve relatively poor results. 
This is because the Gait3D dataset has a serious 3D viewpoint and varying speed, which makes it difficult to extract useful temporal information. 
The SMPLGait achieves relatively the best performance by adding 3D meshes to align the 3D viewpoint.
Meanwhile, we can observe that our method achieves more than 2.4\% Rank-1 accuracy in SMPLGait without 3D meshes. 
This means that in addition to adding additional input data like 3D meshes, better extracting of temporal information is also very important. 

\begin{table}[t]
\centering
\begin{tabular}{l|cccc}
Methods                      			& Rank-1 & Rank-5 & Rank-10 & Rank-20 \\ \midrule[1.5pt]
PoseGait~\cite{pr/LiaoYAH20}	        & 0.23  & 1.05  & 2.23  & 4.28 \\ 
GaitGraph~\cite{corr/abs-2101-11228}	& 1.31  & 3.46 & 5.10  & 7.51 \\  \midrule
GEINet~\cite{icb/ShiragaMMEY16}      	& 6.82  & 13.42 & 16.97  & 21.01 \\
GaitSet~\cite{aaai/ChaoHZF19}		    & 46.28 & 63.58 & 70.26 & 76.82	\\
GaitPart~\cite{cvpr/FanPC0HCHLH20}		& 44.01 & 60.68 & 67.25 & 73.47	\\ 
GaitGL~\cite{Lin_2021_ICCV}	            & 47.28 & 63.56 & 69.32 & 74.18	\\  \midrule
Ours  w/o MTS	                        & 50.42 & 67.89 & 74.28 & 79.38	\\ 
Ours                                    & \textbf{55.32} & \textbf{71.28} & \textbf{76.85} & \textbf{81.55}  \\  \end{tabular} 
\caption{Comparison of the state-of-the-art gait recognition methods on GREW.} 
\vspace{-7mm}
\label{tab:experiments_grew}
\end{table}

\subsection{Experimental results on GREW}
In this section, we compare our method with several state-of-the-art methods on GREW dataset. 
As is shown in Table~\ref{tab:experiments_grew}, PoseGait, GaitGraph, and GEINet get poor results, which is consistent with the result on Gait3D. 
In addition, we find that compared with the performance on Gait3D, the results of GaitGL on GREW was relatively better. 
We think this is because the 3D viewpoint of Gait3D is more serious, resulting in the feature misalignment problem of GaitGL being more obvious. 
Moreover, we can observe that our method achieves the best performance. 
In particular, the RanK-1 accuracy is 8.04\% higher than GaitGL. 
This means that our method has learned more useful temporal information. 




\subsection{Ablation Study}
In this section, we first show that the proposed MTS module can significantly improve the performance of 2D CNN on gait recognition. 
Then, we conduct analytical experiments on several key factors, e.g. sampling strategy, spatial and temporal branches, multiple temporal scales, switch style, and the proportion of switch channels. 
For more details about the computational analysis, please refer to
\textbf{the supplementary material}.

\textbf{Improving 2D CNN baselines}
As is shown in Table~\ref{tab:experiments_gait3d} and Table~\ref{tab:experiments_grew}, it can be seen that by adding our MTS module, there are obvious improvements.~\footnote{Ours w/o MTS is equal to the Baseline in~\cite{zheng2022gait3d}.}
Specifically, the addition of the MTS module resulted in a 5.8\% and 4.9\% improvement in Rank-1 accuracy on Gait3D and GREW, respectively. 
In addition, it should be pointed out that after adding our MTS module, we did not modify the settings of any other hyper-parameters, such as learning rate, iterations, etc., which fully shows the effectiveness of our MTS module. 

\begin{table}[t]
\centering
\begin{tabular}{c|cccc}
Sampling Strategies                     & Rank-1 & Rank-5 & mAP & mINP \\ \midrule[1.5pt]
Uniform Sampling	        & 34.90 & 57.20 & 27.85 & 14.69 \\ 
cyclic Sampling         	& 45.10 & 63.90 & 35.11 & 19.96 \\ 
Non-cyclic Sampling    	    & \textbf{48.70} & \textbf{67.10} & \textbf{37.63} & \textbf{21.92} \\  \end{tabular} 
\caption{Comparison of different sampling strategies.} \vspace{-6mm}
\label{tab:experiments_sampling}
\end{table}

\textbf{Different sampling strategies}
\label{subsubsec:sampling_ablation_study}
We present three sampling strategies for ordered sampling the gait silhouettes, namely uniform sampling, cycle sampling, and non-cycle sampling. 
The uniform sampling strategy refers to the sampling of a gait sequence at equal intervals. 
This method can ensure that each sampling can obtain the complete sequence information as comprehensively as possible. 
However, it is not a good sampling strategy for extracting temporal information, because when the interval is greater than a certain value, the temporal knowledge between frames will be difficult to represent. 
As the results are shown in Table~\ref{tab:experiments_sampling}, its performance is the worst, with Rank-1 10.2\% and 13.8\% lower than that of cycle sampling and non-cycle sampling respectively. 
The cycle sampling strategy is proposed based on the assumption that the gait is a periodic motion. 
To better extract the temporal information, it samples continuous frames as the input of the model. 
When the length of the gait sequence is less than the sampling length, the loop will continue to sample frames from the first frame. 
We carefully analyze the gaits in real-world scenes and find that there are many challenging factors like 3D viewpoint, irregular walking style, occlusion, etc. 
There is often a huge difference between the first and last frames. 
So the assumption of cycle sampling is unsuitable in real-world scenes. 
Based on the above analysis, we propose a new gait data sampling strategy, i.e., Non-cyclic sampling. 
We also load consecutive frames, but when the sequence length is insufficient to the sampling length, the frames are copied from front to back until the sampling length is satisfied, as illustrated in Equ.~\ref{equ:eq_sampling}. 
The results in Table ~\ref{tab:experiments_sampling} show the superiority of the proposed sampling strategy. 

\begin{table}[t]
\centering
\begin{tabular}{cccc|cccc}
Spa. & T.1 & T.2 & T.3                        & Rank-1 & Rank-5 & mAP \\ \midrule[1.5pt]
$\checkmark$ &  &  & 	                                    & 42.90 & 63.90 & 35.19 \\ 
 & $\checkmark$ &  & 	                                    & 39.80 & 60.60 & 31.72 \\ 
$\checkmark$ & $\checkmark$ &  &                            & 45.00 & 66.10 & 35.91 \\ 
$\checkmark$ &  & $\checkmark$ &     	                    & 44.40 & 64.00 & 35.39 \\ 
$\checkmark$ &  &  & $\checkmark$   	                    & 44.20 & 64.60 & 35.12 \\
$\checkmark$ &  & $\checkmark$ & $\checkmark$               & 47.20 & 65.40 & 36.86 \\
$\checkmark$ & $\checkmark$ &  & $\checkmark$               & \textbf{48.70} & \textbf{67.10} & \textbf{37.63} \\  \end{tabular}
\caption{Analysis of spatial and multi-hop temporal switch branches. Spa means the spatial branch, T.1 means the temporal switch 1, and the same analogy applies to T.2 and T.3.} 
\vspace{-6mm}
\label{tab:experiments_spatial_multiple_temporal}
\end{table}

\textbf{Spatial and temporal branches}
We discuss the effects of spatial and temporal branches on gait recognition. 
As is shown in Table~\ref{tab:experiments_spatial_multiple_temporal}, the result of the first row refers to the network containing only the spatial branch, which is our baseline model. 
The second row is the result of using only one temporal switch branch. 
We can see that the results of using only the temporal branch are lower than those of using the space branch. 
In particular, the Rank-1 and mAP accuracy drop 3.1\% and 3.47\%, respectively. 
We think that this is because the temporal switch operation will bring damage to the spatial features. 
The third row is the result of combining the spatial and temporal branches, and we can observe that performance has improved significantly, e.g., the Rank-1 accuracy has increased by 2.1\%. 
Note that in our implementation, the spatial and temporal branches share the same 2D convolution layer, so there is no overhead of model parameters. 

\textbf{Contribution of multiple temporal scales}
In much literature~\cite{mm/Lin_mt3d, aaai/LiZH19}, multi-scale temporal modeling has been studied and its effectiveness has been verified. 
We also explore multiple temporal transfer modeling, and the experimental results are shown in Table~\ref{tab:experiments_spatial_multiple_temporal}. 
First, we can find that by adding the temporal branch of switching one, two, and three strides, the results show a descending trend, for example, the change of Rank-1 is 45.00\% -> 44.40\% -> 44.20\%. 
We think that this is because the temporal information on a large scale is relatively difficult to be learned by the model. 
At the same time, we find that the integration of multiple temporal transfer features can effectively improve the performance of gait recognition, and the best result is that the Rank-1 accuracy reaches 48.70\% when the combination of switching one and three strides simultaneously.

\textbf{Switch style}
There are two styles for switch operation: uni-direction and bi-direction. 
Uni-direction means that only the past frame is fused into the current frame. 
While bi-direction means that both the past and future frames are fused in the current frame, which is illustrated in Fig.~\ref{fig:figure_mtt}. 
As is shown in Table~\ref{tab:experiments_transfer_style}, we can observe that the bi-direction improves the Rank-1 and mAP accuracy by 2.5\% and 1.28\%, respectively, compared to the uni-direction. 
We think this is because the style of bi-direction allows the model to see both past and future information, which helps the model learn more temporal knowledge. 


\begin{figure}[t]
  \centering
   \includegraphics[width=0.95\linewidth]{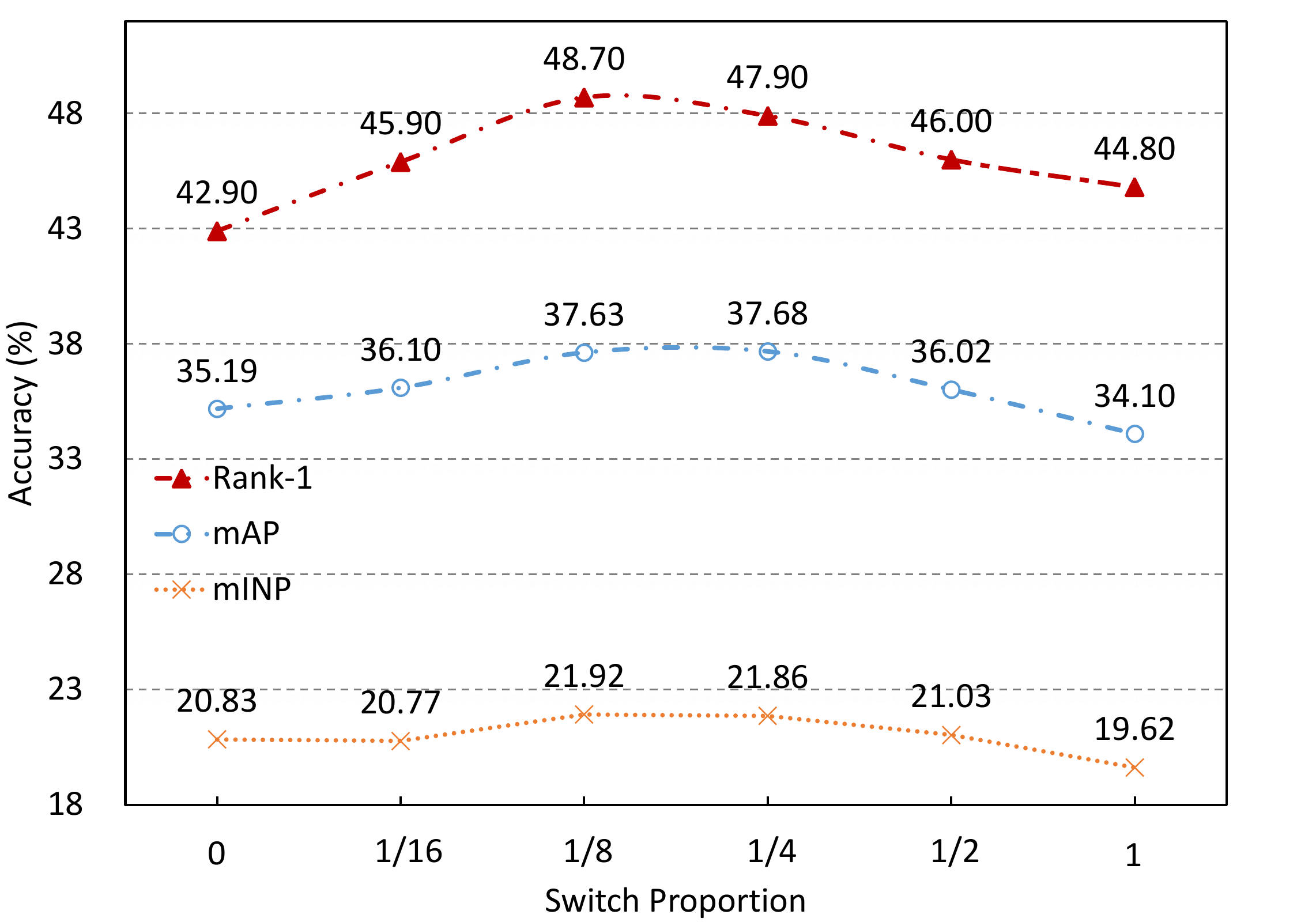}
   \caption{
    Analysis of switch proportion. 
   }
   \label{fig:figure_switch_proportion}
\end{figure}

\begin{table}[t]
\centering
\begin{tabular}{c|cccc}
Switch style                     & Rank-1 & Rank-5 & mAP & mINP \\ \midrule[1.5pt]
Uni-direction         	           & 46.20 & 66.10 & 36.35 & 21.01 \\ 
Bi-direction    	               & \textbf{48.70}	& \textbf{67.10} & \textbf{37.63} & \textbf{21.92} \\  \end{tabular}
\caption{Analysis of switch styles.} 
\vspace{-5mm}
\label{tab:experiments_transfer_style}
\end{table}

\textbf{Proportion of switch channels}
We fix other settings and use different switch proportions, including no switch (2D CNN baseline), partial switch (1/16, 1/8, 1/4, 1/2), and all switch (switch all the channels). 
The results are illustrated in Fig.~\ref{fig:figure_switch_proportion}. 
First, it can be observed that compared with the 2D CNN baseline, the switch operation can effectively improve the accuracy of gait recognition, because the model can learn both spatial and multi-scale temporal information from the MTS module. 
Meanwhile, we find that the performance begins to decline as the switch proportion increases. 
In particular, the performance is even slightly lower than the 2D CNN baseline when transferring all channels, such as the mAP from 35.19\% to 34.10\%. 
We think that switching too many channels will seriously affect the spatial features, thus dragging down the learning efficiency of the model. 

\begin{figure}[t]
  \centering
  \includegraphics[width=0.95\linewidth]{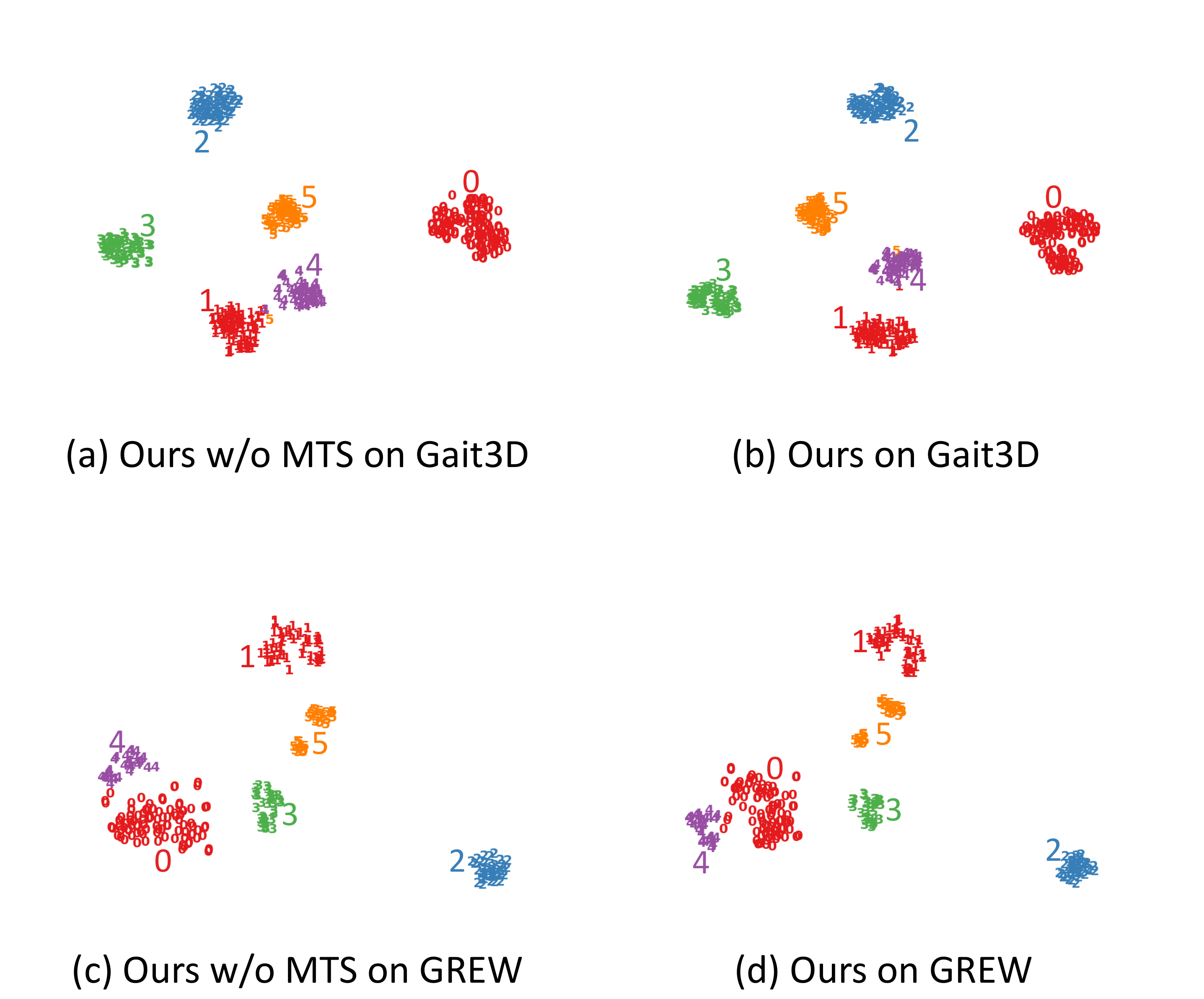}
  \caption{
  Feature distributions are visualized by t-SNE. 
  (a) Our method (w/o MTS) trained on Gait3D. 
  (b) Our method trained on Gait3D. 
  (c) Our method (w/o MTS) trained on GREW. 
  (d) Our method is trained on GREW. 
    (Best viewed in color.)
  }
  \label{fig:figure_tsne}
\end{figure}

\begin{figure}[t]
  \centering
   \includegraphics[width=0.95\linewidth]{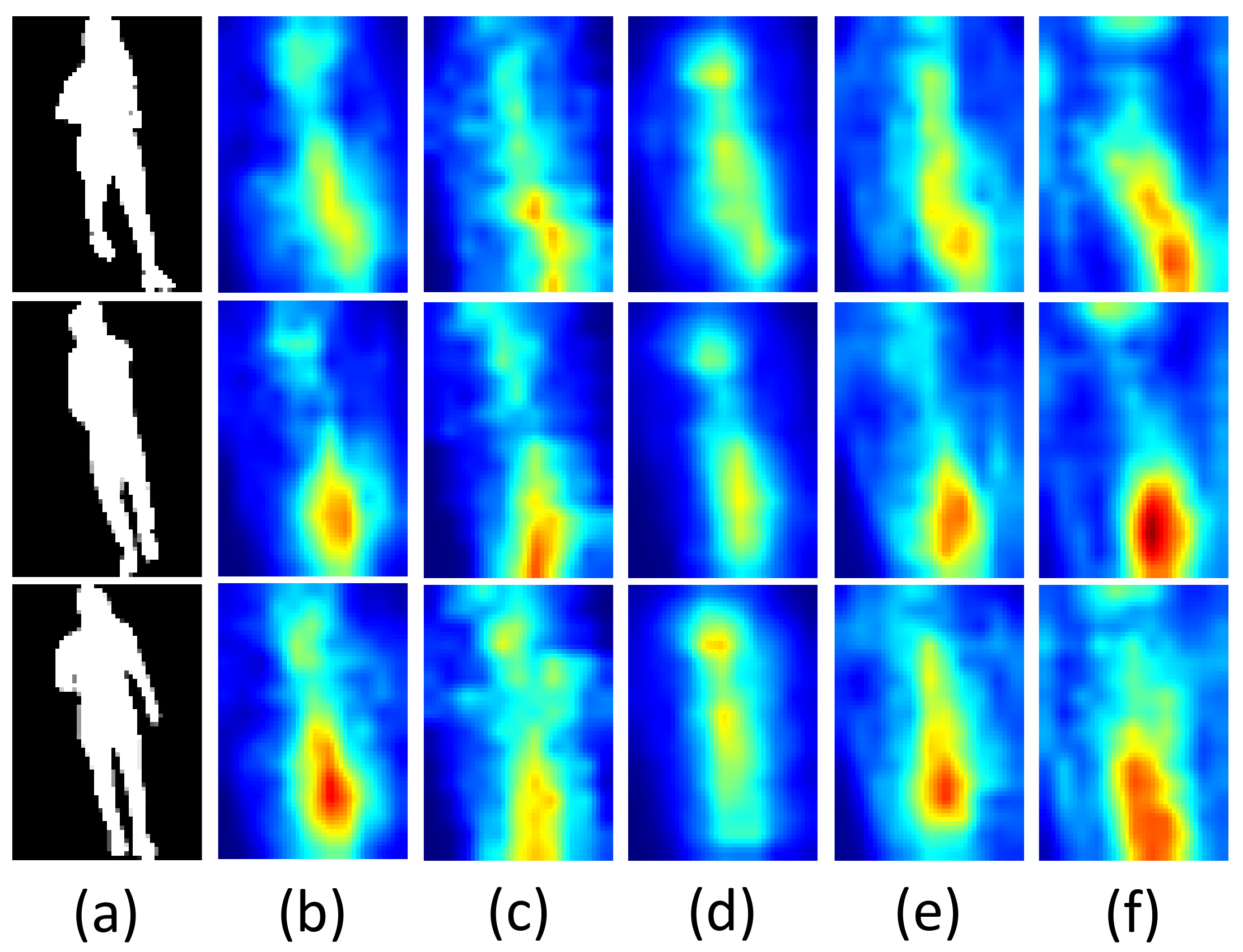}
   \caption{
   Visualization of the original sequence and heatmaps for different methods on Gait3D. (a) input silhouettes. 
   (b) heatmap visualization of GaitSet.
   (c) heatmap visualization of GaitPart.
   (d) heatmap visualization of GLN.
   (e) heatmap visualization of Ours (w/o MTS).
   (f) heatmap visualization of Ours.
    (Best viewed in color.)
   }
   \vspace{-3mm}
   \label{fig:figure_heatmap}
\end{figure}

\subsection{Visualization}

\textbf{Distribution.} 
To observe the discrimination of learned features, we visualize the data distribution of features on Gait3D and GREW datasets. 
As shown in Fig.~\ref{fig:figure_tsne}, with the addition of our MTS module, the intra-class distribution becomes more compact and the distribution of inter-class becomes more distinct. 
In particular, the inter-class distribution of "1" and "4" in Fig.~\ref{fig:figure_tsne} (a) are further separated in Fig.~\ref{fig:figure_tsne} (b). 
the intra-class distribution of "0", "1" and "3" in Fig.~\ref{fig:figure_tsne} (c) are pulled closer in Fig.~\ref{fig:figure_tsne} (d).

\textbf{Heatmap.} 
To better understand the effectiveness of our approach, we also visualize the heatmap figure of several methods, which is shown in Fig.~\ref{fig:figure_heatmap}. 
It can be observed that compared with other methods, our method can make the model pay more attention to the foot movements. 
This shows that the proposed method can indeed promote the learning of temporal information. 

\section{conclusion}
This paper proposes a novel gait recognition network, named MTSGait. 
By combining the spatial branch and multi-hop temporal switch branch, the MTSGait can better learn the spatial-temporal information from the gait sequence. 
In addition, the proposed network is completely based on 2D convolution without the problem of large-scale model parameters and difficulty in training like 3D convolution-based methods. 
Meanwhile, a novel sampling strategy, i.e., Non-cyclic continuous sampling, is proposed, which can improve the model to learn more robust temporal knowledge. 
The experimental results on the public gait in-the-wild datasets demonstrate the effectiveness of our method. 
In the future, we will explore the use of 3D data like the Skinned Multi-Person Linear (SMPL) provided on Gait3D, to further improve the performance of gait recognition in the wild.

\begin{acks}
This work was supported in part by the National Key Research and Development Program of China under Grant 2020YFB1406604, in part by the National Nature Science Foundation of China under Grant 61931008, in part by the Zhejiang Province Nature Science Foundation of China under Grant LZ22F020003, and in part by the HDU-CECDATA Joint Research Center of Big Data Technologies under Grant KYH063120009. 
\end{acks}


\balance
\bibliographystyle{ACM-Reference-Format}
\bibliography{MM2022}

\newpage

\appendix

\begin{table}[t]
\centering
\begin{tabular}{c|cc}
Methods                               & Backbone Parameters   & Inference Time \\ \midrule[1.5pt]
GaitSet~\cite{aaai/ChaoHZF19}	      & 2.59                  & 2.36 \\ 
GaitPart~\cite{cvpr/FanPC0HCHLH20}    & 1.20                  & 4.40 \\ 
GaitGL~\cite{Lin_2021_ICCV}    	      & 2.49                  & 6.75 \\  
Ours w/o MTS      	                  & 3.18                  & 2.20 \\ 
Ours         	                      & 3.18                  & 4.55 \\ \end{tabular} 
\caption{The backbone parameters (M) and inference time (ms/sequence) of
different methods on Gait3D dataset.}
\label{tab:experiments_cpmlex_analysis}
\end{table}

\section{Appendix: The complexity analysis}

In this section, we analyze the complexity of the proposed MTSGait and compare it quantitatively with other methods. 
For simplicity, we analyze it from the perspective of a single convolution layer. 
In addition, we assume that the width and height of the input are equal.

In the single convolution layer, the complexity of 3D convolution is 
\begin{equation}
    O(3DConv) = O(M^{2} \cdot a^{3} \cdot C_{in} \cdot C_{out}),
\label{equ:complex_3dconv}
\end{equation}
where $M=(X-a+2*Padding)/Stride+1$, $X$ is the height or width of the input feature map, $a$ is the length of convolution kernel, $C_{in}$ and $C_{out}$ are the numbers of input and output channels, respectively. 

The complexity of our method is 
\begin{equation}
    O(MTSConv) = O(M^{2} \cdot a^{2} \cdot (k+1) \cdot C_{in} \cdot C_{out}),
\label{equ:complex_mtsconv}
\end{equation}
where $k$ is the number of temporal scale. 
Assume that other parameters are equal, when $(k+1) \leq a$, $O(MTSConv)$ will not exceed $O(3DConv)$. 
The commonly used 2D/3D convolution kernel size is $3 \times 3$ or $5 \times 5$, while we adopt $k =2$ in our implementation.

We also compare our method quantitatively with other methods on Gait3D~\cite{zheng2022gait3d} under the same environment.
The results are shown in Table~\ref{tab:experiments_cpmlex_analysis}, which proves our analysis. 

In summary, our method is comparable with other methods such as GaitSet, GaitPart, and GaitGL, in terms of model parameters and inference time. 
Besides, our accuracy is much better than all the other methods.

\end{document}